\documentclass[review]{elsarticle}

\usepackage{lineno,hyperref}
\usepackage{amsmath}
\usepackage{times}
\usepackage{epsfig}
\usepackage{graphicx}
\usepackage{amsmath}
\usepackage{amssymb}

\usepackage{caption}
\usepackage{algorithm}
\usepackage{algorithmic}
\usepackage{float}
\usepackage{amsmath}
\usepackage{mathrsfs}

\journal{Journal of \LaTeX\ Templates}

\begin{document}

\begin{frontmatter}

\title{Deep graph learning for semi-supervised classification}

\author[mymainaddress]{Guangfeng Lin\corref{mycorrespondingauthor}}
\cortext[mycorrespondingauthor]{Corresponding author}
\ead{lgf78103@xaut.edu.cn}


\author[mymainaddress]{Xiaobing Kang}
\author[mymainaddress]{Kaiyang Liao}
\author[mymainaddress]{Fan Zhao}
\author[mymainaddress]{Yajun Chen}

\address[mymainaddress]{Information Science Department, Xi'an University of Technology,\\
 5 South Jinhua Road, Xi'an, Shaanxi Province 710048, PR China}


\begin{abstract}
Graph learning (GL) can dynamically capture the distribution structure (graph structure) of data based on graph convolutional networks (GCN), and the learning quality of the graph structure directly influences GCN for semi-supervised classification. Existing methods mostly combine the computational layer and the related losses into GCN for exploring the global graph(measuring graph structure from all data samples) or local graph (measuring graph structure from local data samples). Global graph emphasises on the whole structure description of the inter-class data, while local graph trend to the neighborhood structure representation of intra-class data. However, it is difficult to simultaneously balance these graphs of the learning process for semi-supervised classification because of the interdependence of these graphs. To simulate the interdependence, deep graph learning(DGL) is proposed to find the better graph representation for semi-supervised classification. DGL can not only learn the global structure by the previous layer metric computation updating, but also mine the local structure by next layer local weight reassignment. Furthermore, DGL can fuse the different structures by dynamically encoding the interdependence of these structures, and deeply mine the relationship of the different structures by the hierarchical progressive learning for improving the performance of semi-supervised classification. Experiments demonstrate the DGL outperforms state-of-the-art methods on three benchmark datasets (Citeseer,Cora, and Pubmed) for citation networks and two benchmark datasets (MNIST and Cifar10) for images.
\end{abstract}


\end{frontmatter}


\section{Introduction}
Graph ($G(V,E)$, in which $V$ is vertex set for describing dataset and $E$ is edge set for representing the relationship set between data) can capture the relationship of data distribution based on metric method (For example, Euclidean distance,Cosine distance or Kullback-Leibler divergence etc). As a metric representation, graph plays a vital role in pattern recognition. Especially, the recent graph convolutional networks (GCN) have the promising results for many application, for example, human activities \cite{manessi2020dynamic} \cite{zhao2019semantic} \cite{zeng2019graph},facial action unit detection \cite{liu2020relation}, text classification \cite{yao2019graph},and node classification \cite{kipf2016semi} \cite{hu2019hierarchical} \cite{lee2018higher} \cite{qin2018spectral}\cite{lin2019structure}. However, the graph structure is fixed in GCN methods, and it limits GCN for the application of the graph structure loss. Furthermore, the fixed graph structure usually is measured by one metric method, which can not better fit to the distribution of data. Therefore, Graph learning (GL) based on GCN \cite{jiang2019semi} \cite{jiang2019glmnet} \cite{jiang2019unified} \cite{chen2019deep} \cite{hu2019feature} is presented for dynamically mining graph structure of data.

Graph learning faces to a key question, which is the structure relationship learning of data distribution. Existing methods fucus on how to update the graph structure with the metric constraint to optimize the object function \cite{du2020low}\cite{Lin2017Dynamic} or neural networks \cite{jiang2019semi} \cite{jiang2019glmnet} \cite{jiang2019unified} \cite{chen2019deep} \cite{hu2019feature}. The metric constraint usually is defined by two ways. One way is similarity metric learning, which is a global graph structure learning from all data samples. This method often focuses on the difference representation in the inter-class. Another way specifies different weights to different data in it's neighborhood (for example graph attention networks (GAT) \cite{velivckovic2017graph}) for capturing the local graph structure, which tends to the difference description in the intra-class. The global and local graph complement each other for classification. However, existing methods ignores the joint effect of these graphs and the relationship between the global and local graph based on GL for semi-supervised classification. Therefore, DGL is proposed for jointly considering these graphs structure for semi-supervised classification.

\begin{figure*}[ht]
  \begin{center}
\includegraphics[width=1\linewidth]{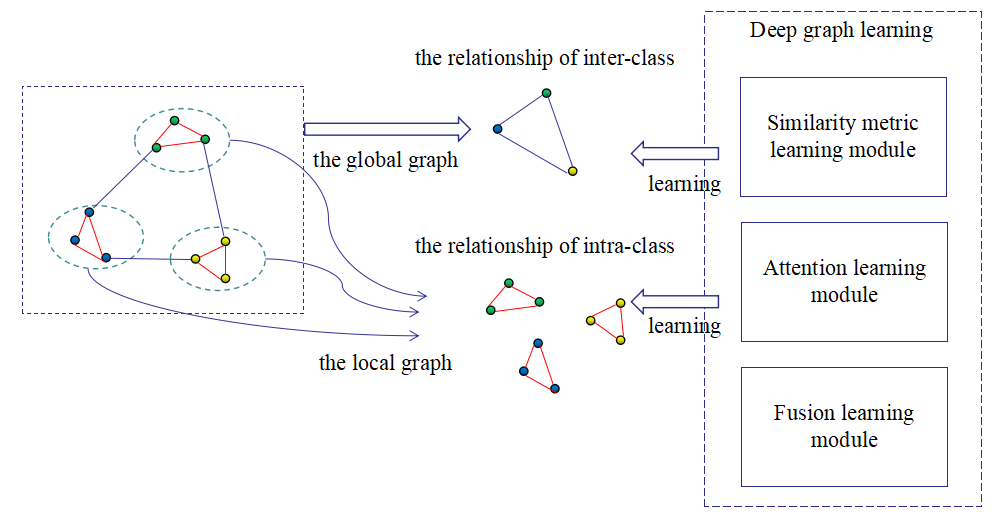}
\end{center}
\vspace{-0.2in}
 \caption{The illustration of deep graph learning for mining the global and local graph.}
  \label{fig-1}
 \end{figure*}

Our main contributions include two points. One is to construct deep graph learning networks for dynamically capturing the global graph by similarity metric learning and local graph by attention learning. Compared with existing methods, the difference of this point focus on the joint consideration the different graphs to further find the distribution structure of the different data. Another is to fuse the global and local graph by the hierarchical progressive learning for semi-supervised classification.In contrast to existing methods, the difference of this point is the dynamic mining the relationship of these graphs to better balance the tendentious contradiction of the different graphs between inter-class and intra-class. the  Figure \ref{fig-1} shows the difference between the global and local graph, and the modules of DGL.

\section{Related Works}
\label{rws}
Graph learning try to automatically construct graph structure from data. Compared with fixed similarity metrics, the difference of GL can dynamically assign the neighbor of each data point, and automatically compute the weight between data points. Therefore, GL can obtain the better accuracy than the fixed graph description by similarity metrics \cite{kang2019robust}.

According to the different learning framework, the recent GL methods can be divided into two categories, which are non-neural networks and neural networks.

One is the methods based on non-neural networks, which attempt to build the optimization function based on the graph generation hypothesis. For example, in terms of completeness hypothesis, self-expressiveness \cite{huang2019auto} \cite{liu2013robust} \cite{kang2018self} regards linear coefficient matrix between data as the graph matrix for the impressive performance in clustering and semi-supervised learning; in accordance with Laplacian graph spectrum, graph learning based on spectral constraints \cite{kumar2019unified} complements the relationship of data by incorporating prior structural knowledge;on the basis of sparse sampling theory, sparse graph learning \cite{hu2019multi} \cite{chen2019adaptive}\cite{pei2019graph} captures few graph connections by adjusting sparsity parameter for improving the classification performance. The superiority of these methods focuses on the relevance between graph generation and constrains, and parameterizes graph generation processing for dynamically controlling graph learning. Because model construction usually be fixed by the specific function, graph structure information from raw data is difficultly mined by iterative boosts.

Another is the approaches based on neural networks, which often simulate the interaction relationship between graph edges and nodes for propagating graph structure information by GCN \cite{jiang2019semi}. In these different networks, these are two types of methods for dynamically computing graph structure. The first type of method is the aggregation of nodes and edges information for updating the weight between nodes layer by layer. For example, hierarchical graph convolutional network(H-GCN) \cite{hu2019hierarchical} repeatedly aggregates similar nodes to hypernodes, and combines one- or two-hop neighborhood information to enlarge the receptive field of each node for encoding graph structure information; edge-labeling graph neural network (EGNN) \cite{kim2019edge} \cite{gong2018exploiting} updates the weight of graph by iteratively aggregating the node representation and the edgelabels with direct exploitation of both intra-cluster similarity and the inter-cluster dissimilarity. The second type of method is the similarity metric of pairwise nodes in some layer. For instance, graph learning-convolutional network(GLCN)\cite{jiang2019semi} optimizes graph structure by learning the transformation relationship of feature difference; dimension-wise separable graph convolution (DSGC) \cite{li2019attributed} uses the relationship among node attributes to complement node relations for representation learning by the covariance metric; graph learning neural networks (GLNNs) \cite{gao2019exploring} iteratively explores the optimization of graphs from both data and tasks by graph Laplacian regularizer; deep iterative and adaptive learning for graph neural networks (DIAL-GNN) \cite{chen2019deep} deals with the graph structure learning problem as a dynamical cosine similarity metric learning problem. These methods mostly consider the global structure from all data sample in the second type of method or the local structure from neighbor data in the first type of method. However, the hierarchical progressive relationship between the global and local graph is ignored.

From above mentions, the methods based on non-neural networks show the better causal relationship between graph structure and the specific optimization function, while the methods based on neural networks demonstrate the stronger learning ability between graph structure and the uncertain optimization networks. It makes the latter be more suitable for further mining the graph structure. Moreover, the similarity metric of pairwise nodes in graph usually directly connect with raw data to easily fit its distribution. Therefore, our proposed method focuses on graph learning based on GCN to find the hierarchical progressive relationship between the global and local graph.

\section{Deep graph learning}

\begin{figure*}[htp]
  \begin{center}
\includegraphics[width=1\linewidth]{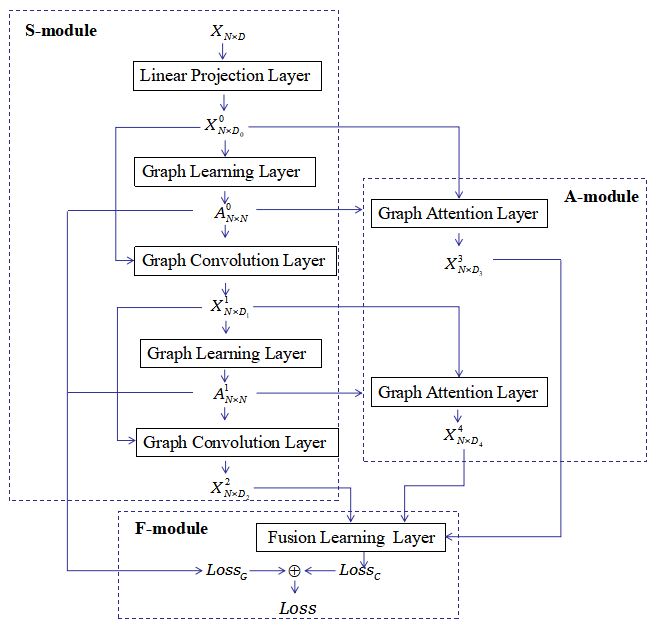}
\end{center}
\vspace{-0.2in}
 \caption{The network frameworks of deep graph learning, which contains similarity metric learning module (S-module), attention learning module(A-module) and fusion learning module(F-module).$X_{N \times D}$,$X_{N \times D_{0}}^{0}$,$X_{N \times D_{1}}^{1}$,$X_{N \times D_{2}}^{2}$,$X_{N \times D_{3}}^{3}$ and $X_{N \times D_{4}}^{4}$ respectively are node representation of each layer(The superscript of the node representation is the serial number of layer, and the subscript of the node representation shows the dimension space of the node representation.); $A_{N\times N}^{0}$ and $A_{N\times N}^{1}$ respectively are the adjacent matrix of the different layer(The superscript of the adjacent matrix is the serial number of layer, and the subscript of the adjacent matrix shows the dimension space of the adjacent matrix.); $Loss_{G}$ is the loss of graph learning; $Loss_{C}$ is the loss of classification; $Loss$ is the total loss of the whole networks.}
  \label{fig-2}
 \end{figure*}

Deep graph learning (DGL) includes three modules, which are similarity metric learning module (S-module), attention learning module(A-module) and fusion learning module(F-module) in figure \ref{fig-2}. Similarity metric learning module implements graph structure computation for dynamically updating global structure relationship based on the raw data or the transformed data. Attention learning module reassigns the weight of the neighbor of each data point for finding the significant local structure based on the global structure. Fusion learning module integrates node representation based on the different graph structure for semi-supervised classification.

\subsection{Similarity metric learning module}
Given data matrix $X\in R^{N\times D}$ ($N$ is the sample number of data, and $D$ is the dimension of each data), Let $X$ be the node representation of graph $G$. We expect to learn $G$ from $X$ for semi-supervised classification. In this module, there are three types of layer for stacking network structure.

\textbf{The first type of layer} is linear projection layer for reducing the dimension of raw data feature. Because the dimension of raw data often leads to the higher computation complexity, the linear transformation of the reduction dimension is expected to implement in this layer.

 \begin{align}
\label{fun-1}
\begin{aligned}
&X^{0}_{N\times D_{0}}=XP,
 \end{aligned}
\end{align}

here, $P\in R^{D \times D_{0}}$ is the linear transformation matrix, and $X^{0}_{N\times D_{0}}$ stands for the output of the linear projection layer.

\textbf{The second type of layer} is graph learning layer for computing the weight of the pairwise nodes. The adjacent relationship $A^{l}_{N\times D_{l}}(i,j)$($i$ and $j$ respectively are the subscript of the different node representation in Graph $G$; $l$ represents the serial number of the layer)can describe this relationship weight, and can be defined as follow.

\begin{align}
\label{fun-2}
\begin{aligned}
A^{l}_{N\times N}(i,j)= &\frac{A(i,j)\exp(ReLU((\alpha^{l})^{T}|x_{i}^{l}-x_{j}^{l}|))}{\sum^{N}_{j}A(i,j)\exp(ReLU((\alpha^{l})^{T}|x_{i}^{l}-x_{j}^{l}|))},
 \end{aligned}
\end{align}

here, $A$ is the normalized adjacent matrix from the initial data source. If $A$ is not available, $A(i,j)=1$.  $ReLU(f)=\max(0,f)$ ($f$ is any variable or matrix) can assure the nonnegativity of $A^{l}_{N\times N}(i,j)$. $x_{i}^{l}\in R^{D_{l} \times 1}$ and $x_{j}^{l}\in R^{D_{l} \times 1}$ respectively are the different row transpose of the input $X^{l}_{N\times D_{l}}$ in the current layer. Equation \ref{fun-2} makes $A^{l}_{N\times N}$ normalized corresponding to its row. $\alpha^{l}\in R^{D_{l} \times 1}$ is weight parameter vector for measuring the significance of the relationship between nodes. Graph learning mainly trains the network for learning $\alpha^{l}$($l$=\{0,1\}).

\textbf{The third type of layer} is graph convolution layer for propagating information based on graph. According to GCN\cite{kipf2016semi}, we can define the graph convolution layer as follow.

\begin{align}
\label{fun-3}
\begin{aligned}
&X_{N \times D_{l+1}}^{l+1}=ReLU(\hat{D^{l}}^{-1/2}\hat{A^{l}}\hat{D^{l}}^{-1/2}X_{N \times D_{l}}^{l}W^{l}),
 \end{aligned}
\end{align}
here, $\hat{A^{l}}=I_{N\times N}+A^{l}_{N\times N}$($I_{N\times N}\in R^{N\times N}$ is the identity matrix); $\hat{D^{l}}(i,i)=\sum_{j}A^{l}_{N\times N}(i,j)$; $W^{l}\in R^{D^{l}\times D^{l+1}}$ is the trainable weight matrix of the current layer.

Similarity metric learning module based on three types layer includes one linear projection layer, two graph learning layer and graph convolution layer from input to output. Especially, two times stack of graph learning layer and graph convolution layer can construct deep network for mining the global graph structure of the different scale node representation.

\subsection{Attention learning module}
In the whole network construction, the global structure generation by similarity metric learning module can initially build local structure information of the neighbor of node representation. However, this local structure information only come from the pair-wise relevance between the current node and all other nodes, but weaken the importance discrimination of the node in the neighborhood of the current node. Therefore, we expect to construct attention learning module by the aggregation of the neighbor information for further capturing the local structure based on the sparse constrains neighborhood of the global structure(we call this process as hierarchical progressive learning).The original GAT \cite{velivckovic2017graph} only can process the binary weight of pair-wise node representation. For example, attention mechanism is built based on node's neighborhood weighted by binary value. However, the weight of the learned graph is real-value, which help to confirm the node's neighborhood with the incorporating the sparse constrains of the global graph structure. Therefore, the operation of attention mechanism is defined as follow.
\begin{align}
\label{fun-4}
\begin{aligned}
&X_{N \times D_{l+1}}^{l+1}=ReLU(\beta^{l}X_{N \times D_{l}}^{l}W^{l}),
 \end{aligned}
\end{align}
here, $\beta^{l}\in R^{N \times N}$ is the attention coefficient matrix, which any entry $\beta^{1}(i,j)$ directly is relevant with $X_{N \times D_{l}}^{l}(i,:)$,$X_{N \times D_{l}}^{l}(j,:)$ and $A_{N\times N}^{l}(i,j)$. Therefore, we define $\beta^{1}(i,j)$ by information aggregation based on graph as follow.
\begin{align}
\label{fun-5}
\begin{aligned}
&\hat{\beta}^{l}(i,j)=\exp{(ReLU(\gamma^{T}[X_{N \times D_{l}}^{l}(i,:)W^{l}\|X_{N \times D_{l}}^{l}(j,:)W^{l}]))}A_{N\times N}^{l}(i,j),
 \end{aligned}
\end{align}
\begin{align}
\label{fun-6}
\begin{aligned}
&\beta^{1}(i,j)=\hat{\beta}^{l}(i,j)/\sum^{N}_{k}\hat{\beta}^{l}(i,k),
 \end{aligned}
\end{align}

here, $\|$ is the concatenation operator for transforming into column vector;$\gamma\in R^{2D_{l+1}\times 1}$ is the aggregation weight, which is shared by the dimension of all pari-wise nodes aggregation.

In attention learning module, we handle the different scale information from the global graph structure by two graph attention layer for further mining local graph structure, which is credible basis for the description of the intra-class.

\subsection{Fusion learning module}
Fusion learning module includes two parts, which respectively are fusion learning layer for the different node representation and loss function for network training propagation.

\textbf{The first part} is fusion learning layer to process the different dimension question of the node representation or the weight balance issue from the different module (similarity metric learning module or attention learning module). From figure \ref{fig-2}, the inputs of this module have $X_{N \times D_{2}}^{2}$ of graph convolution layer output, $X_{N \times D_{3}}^{3}$ and $X_{N \times D_{4}}^{4}$ of the different graph attention layer output. Because this network need deal with classification, we uniform the output dimension of the different module ($D_{2}=D_{3}=D_{4}=C$, $C$ is class number). Therefore, we define fusion learning layer as follow.
\begin{align}
\label{fun-7}
\begin{aligned}
&Z=Softmax(\eta_{1}X_{N \times D_{2}}^{2}+\eta_{2}X_{N \times D_{3}}^{3}+\eta_{3}X_{N \times D_{4}}^{4}),
 \end{aligned}
\end{align}
here $\eta=[\eta_{1},\eta_{2},\eta_{3}]$ is fusion coefficient vector, which encodes the importance of the different node representation.

\textbf{The second part} is loss function definition, which determine the tendency of the network learning. The total loss $Loss$ contains the classification loss $Loss_{C}$ and the graph loss $Loss_{G}$.

In semi-supervised classification, we construct classification loss based on the labeled data by cross-entropy loss for evaluating the error between the predicted label $Z$ and the real label $Y$. Therefore, $Loss_{C}$ is defined as follow.
\begin{align}
\label{fun-8}
\begin{aligned}
&Loss_{C}=-\sum_{k\in S}\sum_{c=1}^{C}Y_{kc}lnZ_{kc},
 \end{aligned}
\end{align}

here,$S$ is the labeled data set; $Y_{kc}$ stands for the $kth$ label data belonging to the $cth$ class; $Z_{kc}$ shows the $kth$ label data predicted as the $cth$ class.

In graph learning, we compute the adjacent matrix $A_{N\times N}^{0}$ and $A_{N\times N}^{1}$ for describing the graph of the different scale. To constrain the properties (sparsity and consistence) of these adjacent matrix, we define the graph loss $Loss_{G}$ as follow.
\begin{align}
\label{fun-9}
\begin{aligned}
Loss_{G}=&\lambda_{1}(X^{T}_{N\times D}(I-A_{N\times N}^{0})X_{N\times D}+X^{T}_{N\times D}(I-A_{N\times N}^{1})X_{N\times D})\\
&+\lambda_{2}(\|A_{N\times N}^{0}\|_{F}^{2}+\|A_{N\times N}^{1}\|_{F}^{2})+\lambda_{3}\|A_{N\times N}^{0}-A_{N\times N}^{1}\|_{F}^{2},
 \end{aligned}
\end{align}
here, the first term can enforce the $X_{N\times D}$ matching with the topology of the graph by graph Laplacian regularizer; the second term can guarantee the sparsity of these adjacent matrixes; the third term can assure the consistence between these adjacent matrixes.

Therefore,the total loss $Loss$ is the sum of $Loss_{C}$ and $Loss_{G}$.
\begin{align}
\label{fun-10}
\begin{aligned}
Loss=Loss_{C}+Loss_{G},
 \end{aligned}
\end{align}

\section{Experiment}
\subsection{Datasets}
For evaluating the proposed DGL method, we carry out experiments in one generated dataset, and six benchmark datasets, which include three the paper-citation networks datasets(Cora,Citeseer and Pubmed\cite{sen2008collective}) and two image datasets(MNIST\cite{lecun1998gradient} and Cifar10\cite{krizhevsky2009learning}).

The synthesized dataset contains 4 classes, each of which has 1000 samples, and includes 4000 samples. These data are randomly synthesized. In experiment, each class samples are divided into four groups, which are 1/100/899, 2/100/898, 3/100/897 and 4/100/896 for training/validation/testing sets. Table \ref{tab-1} show its details.

Cora dataset includes 7 classes that have 2708 grouped publications as nodes represented by one-hot vector in term of the present or absence state of a word in the learned directory and their link relationship graph. Citeseer dataset contains 6 classes that involve 3327 scientific paper described like the same way of Cora dataset and their undirected graph. Pubmed dataset has 3 classes that include 19717 diabetes-related publication indicated by a term frequency-inverse document frequency (TF-IDF)\cite{wu2019comprehensive} and their relevance graph. In these datasets, experiments follow the configuration of the previous work \cite{kipf2016semi}. We select 500 samples for validation and 1000 samples for testing. Table \ref{tab-1} shows the specific information of these datasets.

Cifar10 dataset has 10 classes that consists 50000 natural images\cite{krizhevsky2009learning}. The size of each RGB image is $32\times 32$. We select 10000 images (1000 images for each class) for evaluating the proposed DGL. For representing each image, we use Resnet-20\cite{he2016deep} to extract feature. MNIST dataset contains 10 classes of hand-written digit. We also select 10000 images (1000 images for each class) for assessing the proposed DGL. Each image feature is 784 dimension vector generated by the gray image. Table \ref{tab-1} demonstrates the statistics of these datasets.

\begin{table*}[!ht]
\small
\renewcommand{\arraystretch}{1.0}
\caption{Datasets statistics and the extracted feature in experiments.}
\label{tab-1}
\begin{center}
\newcommand{\tabincell}[2]{\begin{tabular}{@{}#1@{}}#2\end{tabular}}
\begin{tabular}{lp{0.8cm}p{1.0cm}p{1.2cm}p{1.0cm}p{1.5cm}p{1.2cm}p{0.8cm}}
\hline
\bfseries Datasets & \bfseries \tabincell{l}{Classes \\number} & \bfseries \tabincell{l}{Training\\Number} &\bfseries \tabincell{l}{Validating\\Number}& \bfseries \tabincell{l}{Testing\\Number} & \bfseries \tabincell{l}{Total number \\of images} & \bfseries \tabincell{l}{Feature\\dimension}& \bfseries \tabincell{l}{Initial\\graph} \\
\hline \hline
\tabincell{l}{Generated\\ data}&$4$ &$4\sim16$& $400$ &$3596\sim3584$& $4000$ &200&No\\
\hline
Cora &$7$ &$140$& $500$ &$1000$& $2708$ &1433&Yes\\
\hline
Citeseer&$6$ &$120$& $500$ &$1000$& $3327$ &3703&Yes\\
\hline
Pubmed  &$3$ &$59$& $500$ &$1000$& $19717$ &500&Yes\\
\hline
Cifar10 &$10$ &$1000\sim8000$& $1000$ &$8000\sim1000$& $50000$ &128& No\\
\hline
MNIST &$10$ &$1000\sim8000$& $1000$ &$8000\sim1000$& $60000$ &784& No\\
\hline
\end{tabular}
\end{center}
\end{table*}

\subsection{Experimental configuration}
In experiments, we set $D_{0}=70$, $D_{1}=30$ and $D_{2}=D_{3}=D_{4}$, which is equal the classes number. The training maximum episodes of the proposed DGL is $200$. The parameter $\lambda_{1}$, $\lambda_{2}$ and $\lambda_{3}$ respectively are set $0.1$,$0.01$ and $0.001$.  In Cifar10 and MNIST datasets, we select 8 group data for the different training-validating-testing sets(1000-1000-8000, 2000-1000-7000, 3000-1000-6000, 4000-1000-5000, 5000-1000-4000, 6000-1000-3000, 7000-1000-2000 and 8000-1000-1000). In the different datasets, validation set mainly is used for optimizing hyper-parameters, which include the dropout rate for all layer, the number of hidden units and the learning rate.

\subsection{Generated data experiment}
\begin{figure*}[ht]
  \begin{center}
\includegraphics[width=1\linewidth]{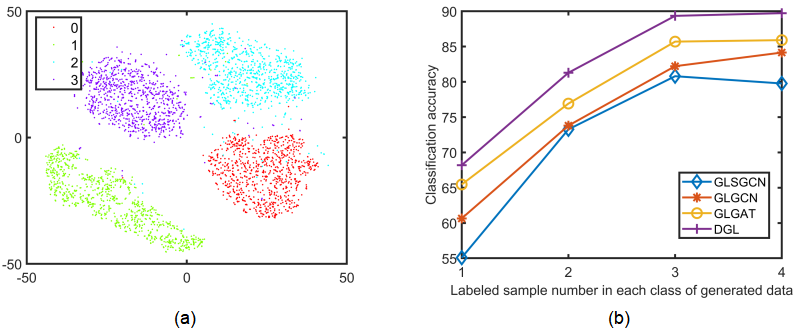}
\end{center}
\vspace{-0.2in}
 \caption{The structure distribution of the two-dimension generated data in (a) and the contrast experiment of the graph learning method in (b).}
  \label{fig-21}
 \end{figure*}

For observing the generated date, we reduce multi-dimension data to two dimension for visualizing data by t-SNE \cite{maaten2008visualizing}. Figure \ref{fig-21} shows the distribution of the generated data in two dimension, and experimental results of four methods, which are the proposed DGL,GLSGCN,GLGCN\cite{jiang2019semi} and GLGAT. GLSGCN and GLGAT is constructed for extending graph learning method in section \ref{extending}. Although the few data are labeled, DGL can still learn the structure distribution of data to obtain the promising results. Therefore, we conduct the following experiments for further evaluating the proposed DGL in the real datasets.

\subsection{Comparison with baseline approaches}
\label{baseline}
In this section, we implement the proposed DGL and the baseline methods, which are GCN\cite{kipf2016semi}, GAT \cite{velivckovic2017graph}, simplifying graph convolutional networks(SGCN)\cite{wu2019simplifying} and GLGCN \cite{jiang2019semi}. GCN can construct the basic architecture of graph representation and classification model by the localized first-order approximation of spectral graph convolutions. GAT can learn the different weights to different nodes in a neighborhood for finding the attentions mechanism of local data. SGCN can eliminate the redundant complexity and computation of GCN by removing nonlinear unit and collapsing operation between the different layer. GLGCN can combine graph learning and graph convolution to optimize the global graph structure. Comparing with these methods, DGL can not only mine the global graph structure by the different scale graph learning layer, but also capture the local graph structure by the different scale graph attention layer. Furthermore, DGL can integrate the node representation from the different graph structure by fusion learning layer. Table \ref{tab-2} shows that DGL has the best performance in these methods. The experimental results between parentheses of GCN, GAT GLGCN come from the literature\cite{jiang2019semi}, while the results of SGCN stem from the literature \cite{wu2019simplifying}.

\begin{table}[!ht]
\small
\renewcommand{\arraystretch}{1.0}
\caption{Comparison of the proposed DGL method with baseline methods (GCN,GAT,SGCN and GLGCN) for semi-supervised classification, average per-class accuracy (\%) is reported based on the same data configurations in the different datasets. The results between parentheses come from the different literatures. All methods use the initial graph for computing model.}
\label{tab-2}
\begin{center}
\newcommand{\tabincell}[2]{\begin{tabular}{@{}#1@{}}#2\end{tabular}}
\begin{tabular}{lp{2.5cm}p{2.5cm}p{2.5cm}}
\hline
\bfseries Method &\bfseries Cora &\bfseries Citeseer &\bfseries Pubmed \\
\hline \hline
GCN\cite{kipf2016semi}  & $81.1\pm0.4 (82.9)$ &$71.0\pm0.2 (70.9)$ & $78.9\pm0.5 (77.9)$ \\
\hline
GAT\cite{velivckovic2017graph}  & $81.4\pm0.8 (83.2)$ &$71.8\pm0.3 (71.0)$ & $78.1\pm0.4 (78.0)$ \\
\hline
SGCN\cite{wu2019simplifying} & $82.3\pm0.5 (81.0)$ &$71.4\pm0.3(71.9)$ & $78.3\pm0.2(78.9)$ \\
\hline
GLGCN\cite{jiang2019semi}  & $82.2\pm0.7(85.5)$ &$72.0\pm0.2(72.0)$ & $78.3\pm0.1(78.3)$ \\
\hline\hline
DGL  & $\textbf{84.8}\pm0.7$ &$\textbf{74.2}\pm0.5$ & $\textbf{80.2}\pm0.2$ \\
\hline
\end{tabular}
\end{center}
\end{table}

\subsection{Comparing with State-of-the-arts}
\label{sota}
Graph learning with neural network shows the promising results for semi-supervised classification. In section \ref{rws}, we summary the graph learning methods based on neural network, find the bias of the global graph structure or the local graph structure in existing methods. Therefore, we try to construct the new graph learning method based on neural network for further mining graph structure and balance the bias of these methods. We compare the proposed DGL with H-GCN \cite{hu2019hierarchical},GLNNs \cite{gao2019exploring},DIAL-GNN\cite{chen2019deep} and GLGCN\cite{jiang2019semi}. The difference of these methods is detailed in section \ref{rws}. Table \ref{tab-3} shows the best performance of the different methods, for example, GLGCN in Cora,and DGL in Citeseer and Pubmed. These methods can obtain the approximate performance in these datasets. For further contrasting the difference between GLGCN and the proposed DGL, we carry out the graph learning experiment in following section.

\begin{table}[!ht]
\small
\renewcommand{\arraystretch}{1.0}
\caption{Comparison of the proposed DGL method with state-of-the-art methods (H-GCN,GLNNs,DIAL-GNN and GLGCN) for semi-supervised classification, average per-class accuracy (\%) is reported based on the same data configurations in the different datasets. The results between parentheses come from the different literatures. All methods use the initial graph for computing model.}
\label{tab-3}
\begin{center}
\newcommand{\tabincell}[2]{\begin{tabular}{@{}#1@{}}#2\end{tabular}}
\begin{tabular}{lp{2.5cm}p{2.5cm}p{2.5cm}}
\hline
\bfseries Method &\bfseries Cora &\bfseries Citeseer &\bfseries Pubmed \\
\hline \hline
H-GCN\cite{hu2019hierarchical}  & $(84.5\pm0.5)$ &$(72.8\pm0.5)$ & $(79.8\pm0.4)$ \\
\hline
GLNNs\cite{gao2019exploring}  & $(83.4)$ &$(72.4)$ & $(76.7)$ \\
\hline
DIAL-GNN\cite{chen2019deep} & $(84.5\pm 0.3)$ &$(74.1\pm0.2)$ & $Null$ \\
\hline
GLGCN\cite{jiang2019semi}  & $\textbf{(85.5)}$ &$(72.0)$ & $(78.3)$ \\
\hline\hline
DGL  & $84.8\pm0.7$ &$\textbf{74.2}\pm0.5$ & $\textbf{80.2}\pm0.2$ \\
\hline
\end{tabular}
\end{center}
\end{table}

\subsection{Comparing with the extended graph learning methods}
\label{extending}
In this section, we involve four methods, which are GLGCN\cite{jiang2019semi}, the proposed DGL and two extended methods (graph learning based on SGCN(GLSGCN) and  graph learning based on GAT(GLGAT)). We use the basic idea of GLGCN to construct GLSGCN and GLGAT. \textbf{GLSGCN} includes a linear projection layer, which reduce the dimension of the original data to $70$,a graph learning layer and the following layers that are same with SGCN\cite{wu2019simplifying}. \textbf{GLGAT} also adds a linear projection layer for reducing the dimension of the data, a graph learning layer and the other layers that have the same configuration like GAT\cite{velivckovic2017graph}. In these experiments, all citation datasets do not use the initial graph, and graph structure can be learned from the original data by the different methods, for instance, GLSGCN and GLGCN tend to capture the global structure; GLGAT shallowly mine the global and local structure; the proposed DGL can deeply consider these structures for semi-supervised classification.

Table \ref{tab-4} demonstrates the performance of the proposed DGL is better than that of other graph learning method. It indicates that deep mining and fusion of the different structure can significantly improve the performance of semi-supervised classification. GLSGCN shows the worse results than other methods in Cora and Citeseer datasets, while this method has the approximate result of other methods in Pubmed datasets. The main reason is that the simplifying structure of GLSGCN has the negative influence for graph structure learning in more categories.

Table \ref{tab-5} shows the experimental results in MINIST image datasets. In the different training sets, DGL can outperform other graph learning methods. The same situation happens in Cifar10 of Table \ref{tab-6}. In all methods, the increasing training data is not a necessary and sufficient condition for the better performance because of the random data selection.
\begin{table}[!ht]
\small
\renewcommand{\arraystretch}{1.0}
\caption{Comparison of the proposed DGL method with the related graph learning methods (GLGCN, GLSGCN, GLGAT and DGL) for semi-supervised classification, average per-class accuracy (\%) is reported based on the same data configurations in the citation datasets (Cora,Citeseer and Pubmed). All methods do not use the initial graph for computing model.}
\label{tab-4}
\begin{center}
\newcommand{\tabincell}[2]{\begin{tabular}{@{}#1@{}}#2\end{tabular}}
\begin{tabular}{lp{2.5cm}p{2.5cm}p{2.5cm}}
\hline
\bfseries Method &\bfseries Cora &\bfseries Citeseer &\bfseries Pubmed \\
\hline \hline
GLSGCN & $55.9\pm0.6$ &$49.6\pm0.3$ & $74.8\pm0.5$ \\
\hline
GLGCN\cite{jiang2019semi}  & $60.1\pm 0.3$ &$64.6\pm 0.2$ & $73.3\pm 0.5$ \\
\hline
GLGAT & $63.1\pm 0.4$ &$65.5\pm0.2$ & $75.3\pm 0.2$ \\
\hline\hline
DGL  & $\textbf{65.3}\pm0.3$ &$\textbf{68.9}\pm0.4$ & $\textbf{76.9}\pm0.5$ \\
\hline
\end{tabular}
\end{center}
\end{table}

\begin{table}[!ht]
\small
\renewcommand{\arraystretch}{1.0}
\caption{Comparison of the proposed DGL method with the related graph learning methods (GLGCN, GLSGCN, GLGAT and DGL) for semi-supervised classification, average per-class accuracy (\%) is reported based on the different data training/validation/testing in the MNIST image datasets.The initial graph for computing model is not available.}
\label{tab-5}
\begin{center}
\newcommand{\tabincell}[2]{\begin{tabular}{@{}#1@{}}#2\end{tabular}}
\begin{tabular}{lp{2.1cm}p{2.1cm}p{2.1cm}p{2.1cm}}
\hline
\bfseries Method &\bfseries \tabincell{l}{MNIST\\1000/1000/8000} &\bfseries \tabincell{l}{MNIST\\2000/1000/7000} &\bfseries \tabincell{l}{MNIST\\3000/1000/6000} &\bfseries\tabincell{l}{MNIST\\4000/1000/5000} \\
\hline \hline
GLSGCN & $37.7\pm0.2$ &$38.7\pm0.4$ & $39.5\pm0.1$  & $39.6\pm0.2$\\
\hline
GLGCN\cite{jiang2019semi}  & $84.9\pm 0.4$ &$85.9\pm 0.2$ & $85.2\pm 0.3$& $88.0\pm 0.2$ \\
\hline
GLGAT & $86.3\pm 0.5$ &$89.9\pm0.2$ & $89.7\pm 0.4$ & $89.2\pm 0.6$\\
\hline\hline
DGL  & $\textbf{89.1}\pm0.6$ &$\textbf{91.4}\pm0.2$ & $\textbf{91.1}\pm0.3$ & $\textbf{92.4}\pm0.5$\\
\hline
\bfseries Method &\bfseries\tabincell{l}{MNIST\\5000/1000/4000} &\bfseries\tabincell{l}{MNIST\\6000/1000/3000} &\bfseries\tabincell{l}{MNIST\\7000/1000/2000} &\bfseries\tabincell{l}{MNIST\\8000/1000/1000} \\
\hline \hline
GLSGCN & $39.4\pm0.3$ &$39.3\pm0.4$ & $38.9\pm0.3$ & $42.7\pm0.5$\\
\hline
GLGCN\cite{jiang2019semi}  & $87.9\pm 0.4$ &$86.4\pm 0.2$ & $88.0\pm 0.5$ & $88.9\pm 0.7$\\
\hline
GLGAT & $89.7\pm 0.3$ &$89.1\pm0.7$ & $89.6\pm 0.4$ & $90.2\pm 0.5$\\
\hline\hline
DGL  & $\textbf{91.1}\pm0.5$ &$\textbf{91.3}\pm0.2$ & $\textbf{91.6}\pm0.6$ & $\textbf{92.4}\pm0.4$\\
\hline
\end{tabular}
\end{center}
\end{table}

\begin{table}[!ht]
\small
\renewcommand{\arraystretch}{1.0}
\caption{Comparison of the proposed DGL method with the related graph learning methods (GLGCN, GLSGCN, GLGAT and DGL) for semi-supervised classification, average per-class accuracy (\%) is reported based on the different data training/validation/testing in the Cifar10 image datasets.The initial graph for computing model is not available.}
\label{tab-6}
\begin{center}
\newcommand{\tabincell}[2]{\begin{tabular}{@{}#1@{}}#2\end{tabular}}
\begin{tabular}{lp{2.1cm}p{2.1cm}p{2.1cm}p{2.1cm}}
\hline
\bfseries Method &\bfseries \tabincell{l}{Cifar10\\1000/1000/8000} &\bfseries \tabincell{l}{Cifar10\\2000/1000/7000} &\bfseries \tabincell{l}{Cifar10\\3000/1000/6000} &\bfseries\tabincell{l}{Cifar10\\4000/1000/5000} \\
\hline \hline
GLSGCN & $63.5\pm0.4$ &$66.4\pm0.3$ & $71.5\pm0.5$  & $72.6\pm0.2$\\
\hline
GLGCN\cite{jiang2019semi}  & $84.2\pm 0.2$ &$79.7\pm 0.5$ & $81.1\pm 0.8$& $86.8\pm 0.4$ \\
\hline
GLGAT & $86.5\pm 0.8$ &$87.4\pm0.5$ & $87.5\pm 0.6$ & $88.0\pm 0.3$\\
\hline\hline
DGL  & $\textbf{87.5}\pm0.5$ &$\textbf{88.8}\pm0.3$ & $\textbf{88.8}\pm0.6$ & $\textbf{88.8}\pm0.4$\\
\hline
\bfseries Method &\bfseries\tabincell{l}{Cifar10\\5000/1000/4000} &\bfseries\tabincell{l}{Cifar10\\6000/1000/3000} &\bfseries\tabincell{l}{Cifar10\\7000/1000/2000} &\bfseries\tabincell{l}{Cifar10\\8000/1000/1000} \\
\hline \hline
GLSGCN & $63.7\pm0.5$ &$73.3\pm0.3$ & $75.5\pm0.6$ & $71.0\pm0.3$\\
\hline
GLGCN\cite{jiang2019semi}  & $83.7\pm 0.9$ &$80.0\pm 0.5$ & $84.5\pm 0.7$ & $80.0\pm 0.7$\\
\hline
GLGAT & $85.2\pm 0.5$ &$86.3\pm0.4$ & $87.5\pm 0.6$ & $87.0\pm 0.3$\\
\hline\hline
DGL  & $\textbf{87.0}\pm0.2$ &$\textbf{88.6}\pm0.5$ & $\textbf{89.0}\pm0.4$ & $\textbf{89.0}\pm0.3$\\
\hline
\end{tabular}
\end{center}
\end{table}

\subsection{Ablation experiments}
\label{ablation}
In this section, we expect to delete some parts form DGL for analyzing the function of the different components. In the proposed DGL, 'deep' has two kinds of meaning. One meaning is the information mining from global structure to local structure (from S-module of DGL to A-module of DGL in figure \ref{fig-2}). Therefore, we delete A-module for simulating the situation of non-local structure, which is called \textbf{DGL-non-local}. Another meaning is the metric learning of the different scale convolution information (two graph learning layers of DGL in \ref{fig-2}). Consequently, we delete the second graph learning layer for imitating the shallow metric learning, which is called \textbf{DGL-shallow-metric}. If DGL  dose not  consider the local graph structure and only care the metric learning of the single layer information, DGL will degrade to GLGCN. So, the intrinsic difference between DGL and GLGCN is the deep graph structure information mining and learning.

Table \ref{tab-7} shows that the performance of DGL is superior to that of other methods. Specially, local graph structure mining by attention mechanism can complement global structure capturing by metric learning, so the performance of DGL-shallow-metric is better than that of GLGCN. Deep metric learning can obtain the more abundant structure information from the different scale node representation, hence the classification accuracy of DGL-non-local outperforms that of GLGCN. The performance of DGL-shallow-metric is obvious better than that of DGL-non-local, and it demonstrates that hierarchical progressive learning from the global structure to the local structure can get the more positive effect than metric learning from the different scale node representation. Furthermore, both factors can be considered for constructing DGL, and DGL can obtain the promising results for semi-supervised classification.

\begin{table}[!ht]
\small
\renewcommand{\arraystretch}{1.0}
\caption{Comparison of the proposed DGL method with GLGCN , and the ablated methods (DGL-non-local and DGL-shallow-metric) for semi-supervised classification, average per-class accuracy (\%) is reported based on the different datasets.The initial graph for computing model is not available.}
\label{tab-7}
\begin{center}
\newcommand{\tabincell}[2]{\begin{tabular}{@{}#1@{}}#2\end{tabular}}
\begin{tabular}{lp{2.5cm}p{2.5cm}p{2.5cm}}
\hline
\bfseries Method &\bfseries Cora &\bfseries Citeseer &\bfseries Pubmed \\
\hline \hline
GLGCN\cite{jiang2019semi}  & $60.1\pm 0.3$ &$64.6\pm 0.2$ & $73.3\pm 0.5$ \\
\hline
DGL-non-local & $62.5\pm0.5$ &$65.9\pm0.2$ & $75.4\pm0.3$ \\
\hline
DGL-shallow-metric & $63.7\pm 0.2$ &$66.2\pm0.5$ & $75.8\pm 0.4$ \\
\hline\hline
DGL  & $\textbf{65.3}\pm0.3$ &$\textbf{68.9}\pm0.4$ & $\textbf{76.9}\pm0.5$ \\
\hline
\bfseries Method &\bfseries \tabincell{l}{MNIST\\1000/1000/8000} &\bfseries\tabincell{l}{Cifar10\\1000/1000/8000} \\
\hline \hline
GLGCN\cite{jiang2019semi}  & $84.9\pm 0.4$ &$84.2\pm 0.2$  \\
\hline
DGL-non-local & $85.7\pm0.3$ &$85.2\pm0.5$ \\
\hline
DGL-shallow-metric & $87.6\pm 0.6$ &$86.9\pm0.3$ \\
\hline\hline
DGL  & $\textbf{89.1}\pm0.6$ &$\textbf{87.5}\pm0.5$ \\
\hline
\end{tabular}
\end{center}
\end{table}

\subsection{Graph learning visualization}
\begin{figure*}[ht]
  \begin{center}
\includegraphics[width=1\linewidth]{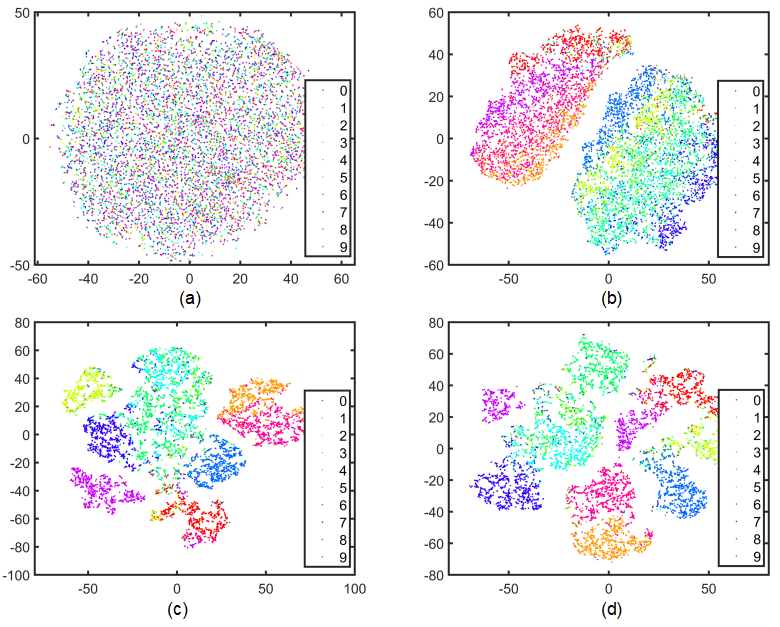}
\end{center}
\vspace{-0.2in}
 \caption{The various structure distribution of the different leaning stage of DGL in Cifar10 dataset. (a) is the structure distribution of episode $1$, (b) for that of episode $50$, (c) for that of episode $100$ and (d) for that of episode $150$. Horizontal and vertical axis respectively stand for the different dimension of data.}
  \label{fig-3}
 \end{figure*}

\begin{figure*}[ht]
  \begin{center}
\includegraphics[width=0.8\linewidth]{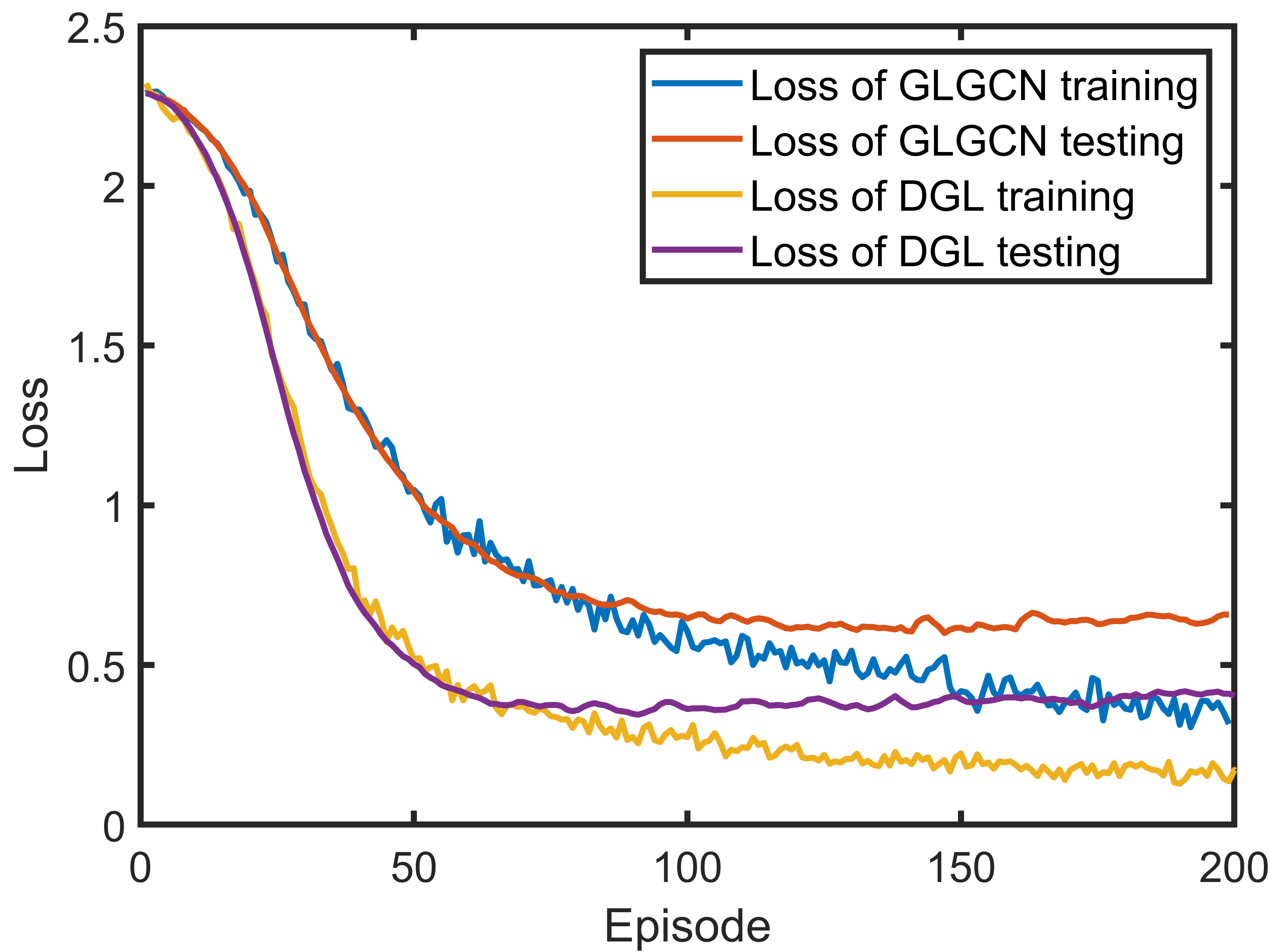}
\end{center}
\vspace{-0.2in}
 \caption{The loss of DGL and GLGCN in training and testing in Cifar10 dataset.}
  \label{fig-4}
 \end{figure*}

For directly observing graph learning process, we reduce multi-dimension node data to two dimension for visualizing data by t-SNE \cite{maaten2008visualizing}. we respectively show the node data distribution of the different episodes(1,50,100,150) in Cifar10 image datasets, in which training/validation/testing data number respectively is set $1000/1000/8000$. Figure \ref{fig-3} shows that the various structure distribution in the different leaning stage. In episode 1, the data distribution presents the hybrid state of the class; in episode 50, the less categories can be separated from all classes; in episode 100, the more categories subsequently can be parted from the whole classes; in episode 150, most of categories can be divided each other. We can observe that the globe and local structure distribution gradually show the aggregation state of the class. Figure \ref{fig-4} indicates that the loss change with episode increasing in DGL and GLGCN. The training or testing loss of DGL obviously is less than that of GLGCN, and it shows that DGL model can obtain the better performance than GLGCN model in training and testing for semi-supervised classification.

\subsection{Experimental results analysis}
\label{analysis}
In experiments, eleven methods are utilized to evaluating the different aspects of the proposed DGL. These method can be divided into four group for the different purpose. The first group includes four baseline methods (GCN\cite{kipf2016semi}, GAT\cite{velivckovic2017graph}, SGCN\cite{wu2019simplifying} and GLGCN \cite{jiang2019semi} in section \ref{baseline}) for cognising the motivation of the proposed DGL. The second group contains four state-of-art methods(H-GCN \cite{hu2019hierarchical},GLNNs \cite{gao2019exploring}, DIAL-GNN\cite{chen2019deep} and GLGCN\cite{jiang2019semi} in section \ref{sota}) for analyzing the advantages and disadvantages between these graph learning methods and the proposed DGL. The third group explores two methods(GLSGCN and GLGAT in section \ref{extending}) based on the main idea of GLGCN \cite{jiang2019semi} for extending the graph learning method based on GCN \cite{kipf2016semi}. The forth group exploits two methods (DGL-non-local and DGL-shallow-metric in section \ref{ablation}) for finding the function of the different components in the proposed DGL. According to the above experiments, we can have the following observations.

\begin{itemize}
\item The performance of DGL outperforms the baseline approaches, which are GCN\cite{kipf2016semi}, GAT \cite{velivckovic2017graph}, SGCN \cite{wu2019simplifying} and GLGCN \cite{jiang2019semi} in section \ref{baseline}.  GCN\cite{kipf2016semi} can reveal the node information propagation based on the statically global graph structure for capturing the data distribution relationship and node representation. GAT \cite{velivckovic2017graph} can assign the weight of the neighborhood in each data node to learn the local graph structure. SGCN \cite{wu2019simplifying} can simplify networks architecture based on the statically global graph structure for reaching the approximating results of GCN. GLGCN \cite{jiang2019semi} can extract the global graph structure from the original data in the networks learning for constructing the basic frameworks of graph learning based on GCN. DGL can not only dynamically mine the global and local graph structure for balancing their effect of the information propagation, but also simultaneously encode the node representation of the different scale outputs for improving the performance of semi-supervised classification.
\item The graph learning methods based GCN (GLGCN\cite{jiang2019semi} and the proposed DGL) have the obvious performance improvement than the non-graph learning methods (GCN\cite{kipf2016semi}, GAT \cite{velivckovic2017graph} and SGCN \cite{wu2019simplifying}). The main reason is that the graph learning methods can dynamically generate graph structure by the parameterized interaction computation, while non-graph learning methods can only depend on the static graph structure in the whole networks learning regardless of the change of each layer. Therefore, the graph learning methods can better fit to the distribution of the transforming data in each layer for enhancing the performance of semi-supervised classification.
\item In the state-of-the-art graph learning methods based on neural networks(H-GCN \cite{hu2019hierarchical},GLNNs \cite{gao2019exploring}, DIAL-GNN\cite{chen2019deep} and GLGCN\cite{jiang2019semi} in section \ref{sota}), the global or local graph structure can be described and mined by hierarchical aggregation or metric learning. The proposed DGL can comprehensively consider the global and local graph structure, and encode their propagation relationship for improving the performance of the networks model. Therefore, DGL can obtain the best performance of Citeseer and Pubmed datasets and the approximated best performance of Cora dataset in these state-of-the-art methods.
\item The extended graph learning methods (GLSGCN and GLGAT in section \ref{extending}) conbine the main idea of GLGCN\cite{jiang2019semi} with GAT \cite{velivckovic2017graph} or SGCN \cite{wu2019simplifying} for finding the adaptation of the graph learning method. GLSGCN can get the worse performance than GLGCN \cite{jiang2019semi},while GLGAT can obtain the better performance than GLGCN \cite{jiang2019semi}. It shows that nonlinear unit layer have the stronger learning ability for dynamically generating graph structure. DGL outperforms GLSGCN and GLGAT, and it demonstrates that the different scale metric learning (from the global to the local graph structure and from the different layers) can contribute to the construction of the graph learning model.
\item The proposed DGL method can delete the different components to formulate the different ablation methods (DGL-non-local and DGL-shallow-metric in section \ref{ablation}). DGL-non-local method emphasises on the global graph structure learning from the different scale node representation, while DGL-shallow-metric focuses on the balance learning between the global and the local graph structure in single layer. The performance of DGL-shallow-metric is superior to that of DGL-non-local, and it indicates that the depth mining from the global graph structure to the local graph structure has the more obvious effect than deep metric learning from the different scale outputs. However, two factors is simultaneously considered to build DGL that can obtain the promising results for semi-supervised classification.
\item In the extended graph learning experiment, the different graph learning method shows the approximate results with training/validation/testing change. It reveals that graph learning process can complement the insufficient number of the training samples for improving the generalization of the model. Therefore, in Table \ref{tab-5} and \ref{tab-6}, this situation happens in the experimental results of the different graph learning methods.
\end{itemize}

\section{Conclusion}
We have presented deep graph learning (DGL) method to address the global and local graph integration learning for improving semi-supervised classification. The proposed DGL can not only use graph learning layer and graph attention layer for hierarchical progressive graph structure mining, but also adopt two graph learning layers for deep capturing the global graph structure information from the different scale node representation. Furthermore, DGL can balance the difference between the global and local graph structure for finding the abundant data relationship, and fusion the node representation of the different layers for enhancing semi-supervised classification. Finally, DGL can automatically generate graph structure in networks learning, and dynamically encode the various information of the different layers. Experimental results and analysis shows that the proposed DGL method is promising for node classification on Citeseer,Cora, Pubmed,MNIST and Cifar10 datasets.

\section{Acknowledgements}
The authors would like to thank the anonymous reviewers for their insightful comments that help improve the quality of this paper. This work was supported by NSFC (Program No.61771386,Program No.61671376 and Program No.61671374).

\bibliography{mybibfile}

\begin{thebibliography}{10}
\expandafter\ifx\csname url\endcsname\relax
  \def\url#1{\texttt{#1}}\fi
\expandafter\ifx\csname urlprefix\endcsname\relax\def\urlprefix{URL }\fi
\expandafter\ifx\csname href\endcsname\relax
  \def\href#1#2{#2} \def\path#1{#1}\fi

\bibitem{manessi2020dynamic}
F.~Manessi, A.~Rozza, M.~Manzo, Dynamic graph convolutional networks, Pattern
  Recognition 97 (2020) 107000.

\bibitem{zhao2019semantic}
L.~Zhao, X.~Peng, Y.~Tian, M.~Kapadia, D.~N. Metaxas, Semantic graph
  convolutional networks for 3d human pose regression, in: Proceedings of the
  IEEE Conference on Computer Vision and Pattern Recognition, 2019, pp.
  3425--3435.

\bibitem{zeng2019graph}
R.~Zeng, W.~Huang, M.~Tan, Y.~Rong, P.~Zhao, J.~Huang, C.~Gan, Graph
  convolutional networks for temporal action localization, in: Proceedings of
  the IEEE International Conference on Computer Vision, 2019, pp. 7094--7103.

\bibitem{liu2020relation}
Z.~Liu, J.~Dong, C.~Zhang, L.~Wang, J.~Dang, Relation modeling with graph
  convolutional networks for facial action unit detection, in: International
  Conference on Multimedia Modeling, Springer, 2020, pp. 489--501.

\bibitem{yao2019graph}
L.~Yao, C.~Mao, Y.~Luo, Graph convolutional networks for text classification,
  in: Proceedings of the AAAI Conference on Artificial Intelligence, Vol.~33,
  2019, pp. 7370--7377.

\bibitem{kipf2016semi}
T.~N. Kipf, M.~Welling, Semi-supervised classification with graph convolutional
  networks, arXiv preprint arXiv:1609.02907 (2016).

\bibitem{hu2019hierarchical}
F.~Hu, Y.~Zhu, S.~Wu, L.~Wang, T.~Tan, Hierarchical graph convolutional
  networks for semi-supervised node classification, arXiv preprint
  arXiv:1902.06667 (2019).

\bibitem{lee2018higher}
J.~B. Lee, R.~A. Rossi, X.~Kong, S.~Kim, E.~Koh, A.~Rao, Higher-order graph
  convolutional networks, arXiv preprint arXiv:1809.07697 (2018).

\bibitem{qin2018spectral}
A.~Qin, Z.~Shang, J.~Tian, Y.~Wang, T.~Zhang, Y.~Y. Tang, Spectral--spatial
  graph convolutional networks for semisupervised hyperspectral image
  classification, IEEE Geoscience and Remote Sensing Letters 16~(2) (2018)
  241--245.

\bibitem{lin2019structure}
G.~Lin, J.~Wang, K.~Liao, F.~Zhao, W.~Chen, Structure fusion based on graph
  convolutional networks for semi-supervised classification, arXiv preprint
  arXiv:1907.02586 (2019).

\bibitem{jiang2019semi}
B.~Jiang, Z.~Zhang, D.~Lin, J.~Tang, B.~Luo, Semi-supervised learning with
  graph learning-convolutional networks, in: Proceedings of the IEEE Conference
  on Computer Vision and Pattern Recognition, 2019, pp. 11313--11320.

\bibitem{jiang2019glmnet}
B.~Jiang, P.~Sun, J.~Tang, B.~Luo, Glmnet: Graph learning-matching networks for
  feature matching, arXiv preprint arXiv:1911.07681 (2019).

\bibitem{jiang2019unified}
B.~Jiang, X.~Jiang, A.~Zhou, J.~Tang, B.~Luo, A unified multiple graph learning
  and convolutional network model for co-saliency estimation, in: Proceedings
  of the 27th ACM International Conference on Multimedia, ACM, 2019, pp.
  1375--1382.

\bibitem{chen2019deep}
Y.~Chen, L.~Wu, M.~J. Zaki, Deep iterative and adaptive learning for graph
  neural networks, arXiv preprint arXiv:1912.07832 (2019).

\bibitem{hu2019feature}
W.~Hu, X.~Gao, G.~Cheung, Z.~Guo, Feature graph learning for 3d point cloud
  denoising, arXiv preprint arXiv:1907.09138 (2019).

\bibitem{du2020low}
H.~Du, L.~Ma, G.~Li, S.~Wang, Low-rank graph preserving discriminative
  dictionary learning for image recognition, Knowledge-Based Systems 187 (2020)
  104823.

\bibitem{Lin2017Dynamic}
G.~Lin, K.~Liao, B.~Sun, Y.~Chen, F.~Zhao, Dynamic graph fusion label
  propagation for semi-supervised multi-modality classification, Pattern
  Recognition 68 (2017) 14--23.

\bibitem{velivckovic2017graph}
P.~Veli{\v{c}}kovi{\'c}, G.~Cucurull, A.~Casanova, A.~Romero, P.~Lio,
  Y.~Bengio, Graph attention networks, arXiv preprint arXiv:1710.10903 (2017).

\bibitem{kang2019robust}
Z.~Kang, H.~Pan, S.~C. Hoi, Z.~Xu, Robust graph learning from noisy data, IEEE
  transactions on cybernetics (2019).
\newblock \href {https://doi.org/10.1109/TCYB.2018.2887094}
  {\path{doi:10.1109/TCYB.2018.2887094}}.

\bibitem{huang2019auto}
S.~Huang, Z.~Kang, I.~W. Tsang, Z.~Xu, Auto-weighted multi-view clustering via
  kernelized graph learning, Pattern Recognition 88 (2019) 174--184.

\bibitem{liu2013robust}
G.~Liu, Z.~Lin, S.~Yan, J.~Sun, Y.~Yu, Y.~Ma, Robust recovery of subspace
  structures by low-rank representation, IEEE transactions on pattern analysis
  and machine intelligence 35~(1) (2013) 171--184.

\bibitem{kang2018self}
Z.~Kang, X.~Lu, J.~Yi, Z.~Xu, Self-weighted multiple kernel learning for
  graph-based clustering and semi-supervised classification, arXiv preprint
  arXiv:1806.07697 (2018).

\bibitem{kumar2019unified}
S.~Kumar, J.~Ying, J.~V. d.~M. Cardoso, D.~Palomar, A unified framework for
  structured graph learning via spectral constraints, arXiv preprint
  arXiv:1904.09792 (2019).

\bibitem{hu2019multi}
Z.~Hu, F.~Nie, W.~Chang, S.~Hao, R.~Wang, X.~Li, Multi-view spectral clustering
  via sparse graph learning, Neurocomputing (2019).
\newblock \href {https://doi.org/10.1016/j.neucom.2019.12.004}
  {\path{doi:10.1016/j.neucom.2019.12.004}}.

\bibitem{chen2019adaptive}
P.~Chen, L.~Jiao, F.~Liu, Z.~Zhao, J.~Zhao, Adaptive sparse graph learning
  based dimensionality reduction for classification, Applied Soft Computing
  (2019).
\newblock \href {https://doi.org/10.1016/j.asoc.2019.04.029}
  {\path{doi:10.1016/j.asoc.2019.04.029}}.

\bibitem{pei2019graph}
X.~Pei, J.~Zou, W.~Chen, Graph learning via edge constrained sparse
  representation for image analysis, IEEE Access 7 (2019) 42408--42417.

\bibitem{kim2019edge}
J.~Kim, T.~Kim, S.~Kim, C.~D. Yoo, Edge-labeling graph neural network for
  few-shot learning, in: Proceedings of the IEEE Conference on Computer Vision
  and Pattern Recognition, 2019, pp. 11--20.

\bibitem{gong2018exploiting}
L.~Gong, Q.~Cheng, Adaptive edge features guided graph attention networks,
  arXiv preprint arXiv:1809.02709 (2018).

\bibitem{li2019attributed}
Q.~Li, X.~Zhang, H.~Liu, X.-M. Wu, Attributed graph learning with 2-d graph
  convolution, arXiv preprint arXiv:1909.12038 (2019).

\bibitem{gao2019exploring}
X.~Gao, W.~Hu, Z.~Guo, Exploring structure-adaptive graph learning for robust
  semi-supervised classification, arXiv preprint arXiv:1904.10146 (2019).

\bibitem{sen2008collective}
P.~Sen, G.~Namata, M.~Bilgic, L.~Getoor, B.~Galligher, T.~Eliassi-Rad,
  Collective classification in network data, AI magazine 29~(3) (2008) 93--93.

\bibitem{lecun1998gradient}
Y.~LeCun, L.~Bottou, Y.~Bengio, P.~Haffner, et~al., Gradient-based learning
  applied to document recognition, Proceedings of the IEEE 86~(11) (1998)
  2278--2324.

\bibitem{krizhevsky2009learning}
A.~Krizhevsky, G.~Hinton, et~al., Learning multiple layers of features from
  tiny images, Tech. rep., Citeseer (2009).

\bibitem{wu2019comprehensive}
Z.~Wu, S.~Pan, F.~Chen, G.~Long, C.~Zhang, P.~S. Yu, A comprehensive survey on
  graph neural networks, arXiv preprint arXiv:1901.00596 (2019).

\bibitem{he2016deep}
K.~He, X.~Zhang, S.~Ren, J.~Sun, Deep residual learning for image recognition,
  in: Proceedings of the IEEE conference on computer vision and pattern
  recognition, 2016, pp. 770--778.

\bibitem{maaten2008visualizing}
L.~v.~d. Maaten, G.~Hinton, Visualizing data using t-sne, Journal of machine
  learning research 9~(Nov) (2008) 2579--2605.

\bibitem{wu2019simplifying}
F.~Wu, T.~Zhang, A.~H.~d. Souza~Jr, C.~Fifty, T.~Yu, K.~Q. Weinberger,
  Simplifying graph convolutional networks, arXiv preprint arXiv:1902.07153
  (2019).

\end{thebibliography}

\end{document}